\documentclass[british,letterpaper]{article}
\usepackage[latin9]{inputenc}
\usepackage{xcolor}
\usepackage{amsmath}
\usepackage{graphicx}

\makeatletter
\usepackage{aaai23}  
\usepackage{times}  
\usepackage{helvet}  
\usepackage{courier}  
\usepackage[hyphens]{url}  
\usepackage{graphicx} 
\usepackage{natbib}  
\usepackage{caption} 
\usepackage{algorithm}
\usepackage{algorithmic}

\usepackage{newfloat}
\usepackage{listings}
\DeclareCaptionStyle{ruled}{labelfont=normalfont,labelsep=colon,strut=off} 
\floatstyle{ruled}
\newfloat{listing}{tb}{lst}{}
\floatname{listing}{Listing}
\title{Memory-Augmented Theory of Mind Network}
\author{
    Dung Nguyen, 
    Phuoc Nguyen, 
    Hung Le, 
    Kien Do, 
    Svetha Venkatesh, 
    Truyen Tran 
}
\affiliations{
    Applied Artificial Intelligence Institute (A$^2$I$^2$), Deakin University, Geelong, Australia \\
    \{dung.nguyen,phuoc.nguyen,thai.le,k.do,svetha.venkatesh,truyen.tran\}@deakin.edu.au

}

\usepackage{soul}
\usepackage{url}
\usepackage{inputenc}
\usepackage{amsmath}
\usepackage{amsthm}
\usepackage{booktabs}
\usepackage{algorithm}
\usepackage{algorithmic}
\usepackage{microtype}
\@ifundefined{showcaptionsetup}{}{%
 \PassOptionsToPackage{caption=false}{subfig}}
\usepackage{subfig}
\makeatother

\usepackage{babel}
\begin{document}
\global\long\def\Model{\mathtt{ToMMY}}%

\title{Memory-Augmented Theory of Mind Network}
\maketitle
\begin{abstract}
Social reasoning necessitates the capacity of theory of mind (ToM),
the ability to contextualise and attribute mental states to others
without having access to their internal cognitive structure. Recent
machine learning approaches to ToM have demonstrated that we can train
the observer to read the past and present behaviours of other agents
and infer their beliefs (including false beliefs about things that
no longer exist), goals, intentions and future actions. The challenges
arise when the behavioural space is complex, demanding skilful space
navigation for rapidly changing contexts for an extended period. We
tackle the challenges by equipping the observer with novel neural
memory mechanisms to encode, and hierarchical attention to selectively
retrieve  information about others. The memories allow rapid, selective
querying of distal related past behaviours of others to deliberatively
reason about their current mental state, beliefs and future behaviours.
This results in $\Model$, a theory of mind model that learns to reason
while making little assumptions about the underlying mental processes.
We also construct a new suite of experiments to demonstrate that memories
facilitate the learning process and achieve better theory of mind
performance, especially for high-demand false-belief tasks that require
inferring through multiple steps of changes.

\end{abstract}

\section{Introduction}

Human social interactions necessitate a skill known as theory of mind
(ToM) to infer the mental states of others without having access to
their latent characteristics, internal states and computation processes.
Instead, we can rely on social cues and past behaviours to construct
\emph{models} of others, thereby attributing mental states to them,
for example, inferring their beliefs and intentions. The models need
not perfectly match with the true hidden internal mental states but
facilitate accurate social prediction and planning \cite{premack1978does,gallese1998mirror,rusch2020theory,langley2022theory}. 

Since often we can only have access to others' past behaviours and
current observable context, it is plausible that we need memory to
store and represent the past of others, to contextualise the present,
to draw analogies between the present and the related past, and to
reason about possibilities \cite{grant2017can}. Cognitive scientists
have employed memory of structured representation of tasks to enable
analogical reasoning in ToM, for example, to recognise false beliefs
\cite{rabkina2017towards}. Likewise, the work of \cite{nguyen2021theory}
uses instance matching to model the human's ToM ability. This cognitive
model assumes that the observer, who is constructing a model of the
actor, also has access to rewards for each experience of the actor.
These works rely on either domain knowledge to construct the task
structure or information that can be inaccessible to the observer.

In this paper, we take an alternative road to\emph{ learn the memory
mechanisms} to build the computational ToM capability into artificial
social agents. Here mentalising and predicting behaviours of an actor
in a partially observable environment are treated as a task to be
learnt \cite{rabinowitz2018machine}. In the learning phase, the ToM
observer first acquires a general prior mental model from the observed
behavioural episodes of training actors. In the execution phase, upon
seeing an actor and its partial episode, the observer rapidly updates
the specific posterior about the actor. Realising this strategy, we
equip the observer with a new memory-augmented ToM network dubbed
$\Model$ (\textbf{T}heory \textbf{o}f \textbf{M}ind with \textbf{M}emor\textbf{Y}).
Central to this architecture is the memory module, arranged as key-value
pairs storing the past behaviours of the actor. The analogy-making
capability is learned from training data, and once trained, it works
by selectively querying relevant memory keys for a given context to
retrieve corresponding predictive values. The retrieved values, combined
with the selective few states of the current episode, constitute the
posterior of the actor's mental state, which serves as an input for
predicting its future behaviours. Learning memory-augmented neural
networks is a powerful technique for multi-step reasoning \cite{sukhbaatar2015end,graves2016hybrid},
handling rare events \cite{kaiser2017learning}, meta-learning \cite{santoro2016meta}
and rapid reinforcement learning \cite{le2021model}. However, little
work has been done in the area of mentalising other agents in social
settings. 

\begin{figure*}
\begin{centering}
\includegraphics[width=0.9\textwidth]{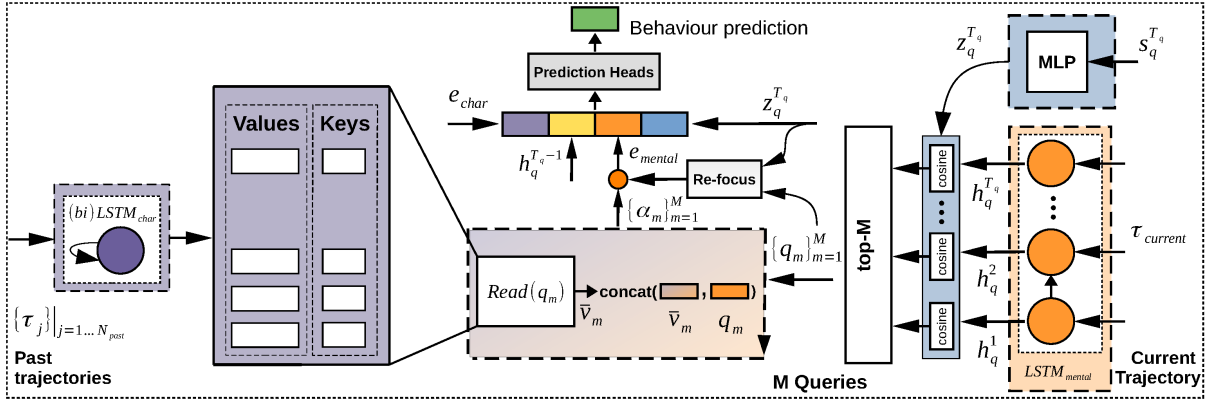}
\par\end{centering}
\caption{\label{fig_mem_tom_working_sparse} Architecture of memory-based theory
of mind agent ($\protect\Model$). Past trajectories of the actor
are encoded into memory of (key, value) pairs. A selective set of
events in current trajectory are embedded into mental states, which
used to query the memory. The prediction heads then generate future
prediction using the retrieved information, the current mental state
and the world state.}
\end{figure*}

To assess the performance of $\Model$ we introduce a new \emph{false-belief}
test to evaluate the ToM ability under high reasoning demand. False-belief
tasks determine if the actor is maintaining an outdated belief about
something that no longer holds. A classic example is the Sally-Anne
Test \cite{wimmer1983beliefs,baron1985does}, in which Anne secretly
moves a toy out of the original box, causing Sally to falsely believe
that the toy is still there. A version of the Sally-Anne Test for
testing artificial theory of mind agents introduced in \cite{rabinowitz2018machine,nguyen2021theory}
take the form of grid-worlds. When the actor tries to achieve the
sub-task before reaching a goal, the position of the goal will be
changed, a so-called \emph{swap event}. This event induces a false
belief in the actor, and the observer needs to take the actor's perspective
to understand the actor's false belief. This test is simple to some
extent, requiring a low-demand mental capacity of social reasoning
to pass. In our proposal, there are multiple steps in the trajectory
of the agent with hidden information; hence the ToM observer needs
to keep track of these steps in order to infer the actor's belief.
More concretely, consider a robot helper: It knows more than the human
in a particular area; however, it cannot observe all events in other
areas where both (the robot and human) are not currently present.
To pass our test, the ToM agents must refer to the states in the last
visits to areas to understand whether the actor has a false belief
and combine it with past behaviours to predict future behaviours.
Therefore, in this high-demand false-belief task, the memory mechanism
is the key to achievement. This matches with the prediction of developmental
psychology that theory of mind in later development requires memory
when humans need to flexibly execute belief reasoning and deal with
a complex situation \cite{apperly2009humans}.

\subsection{Related Work}

The $\Model$ agent makes little assumption about the underlying mental
structure of others. This differs from cognitive science works such
as those in \cite{baker2011bayesian,baker2017rational}, hypothesising
that human will maximise their own utility when mentalising about
others. In AI, ToM is traditionally studied in plan recognition \cite{geib2009probabilistic,sohrabi2016plan},
assuming structural knowledge of the domain. Recent works have been
studied in the NLP domain \cite{nematzadeh2018evaluating,le2019revisiting,arodi2021textual},
focusing on second-order false-belief tasks. Instead, we take a \emph{meta-learning}
stand, similar to that in \cite{rabinowitz2018machine}, updating
the posterior of latent characters for unseen actors, starting from
the prior learnt from seen actors. This work compresses the entire
history of an actor into a vector representing its character, which
is then combined with the current episode and state to predict goal,
intention, action, and successor representations. Such compression
rapidly forgets information from the far past, making it difficult
to reason about rare and distantly separated situations. Our memory
mechanisms effectively tackle this forgetting problem.

There are various methods to measure the ToM ability of humans in
developmental psychology \cite{beaudoin2020systematic}, e.g. either
direct assessment as Sally-Anne Test or indirect assessment via the
violation of expectation (VoE) \cite{onishi200515}, anticipatory
looking \cite{clements1994implicit} and active helping \cite{knudsen201218}.
In \cite{gandhi2021baby} and \cite{shu2021agent}, the authors used
the VoE to evaluate the ToM ability. \cite{rabinowitz2018machine,nguyen2021theory}
constructed a test to mimic the Sally-Anne Test in artificial intelligence. 

Our false-belief testbed is designed for more complex situations that
demand long-term memory. Measuring the cognitive load of a task is
an active field in cognitive science \cite{zheng2017cognitive}. Here,
we intuitively add distractors to increase the cognitive load of the
false-belief task. Measuring the difficulty of the task is a direction
that requires investigation as it is necessary for assessing the ability
of artificial agents.

\section{Problem Formulation}

We study the general setting under the partially observable Markov
decision processes (POMDPs) framework. The \emph{observer} or \emph{theory
of mind (ToM) agent} first observes a set of $N_{past}$ \emph{past
trajectories} $\left\{ \tau_{j}\right\} $ of an \emph{actor} in multiple
environments for $j=1,\dots,N_{past}$. Each past trajectory $\tau_{j}$
is a sequence of state-action pairs $\left(s_{j}^{t},a_{j}^{t}\right)$
for $t=1,2,...,T_{j}$. Upon seeing the current trajectory $\tau_{q}$
which is a sequence of $\left(s_{q}^{t},a_{q}^{t}\right)$ from $t=1$
up to time $T_{q}-1$, and the current state $s_{q}^{T_{q}}$ at time
$T_{q}$, the ToM agent will predict the goal (or preference) and
the future behaviours of the actor, including the intention, the next
action, and the future visit states of the actor (via successor representations
\cite{dayan1993improving}).

\section{Memory-Augmented Theory of Mind Network}

We now describe $\Model$ (\textbf{T}heory \textbf{o}f \textbf{M}ind
with \textbf{M}emor\textbf{Y}), our theory of mind (ToM) agent, who
observes an actor acting in a partially observable environment. $\Model$
is implemented as a memory-augmented neural network as illustrated
in Fig.~\ref{fig_mem_tom_working_sparse}. $\Model$ maintains an
episodic memory of the past trajectories of the actor, which consist
of information reflecting the general latent character of the actor.
From the current incomplete trajectory, $\Model$ selectively chooses
several events to serve as queries to retrieve relevant past events
from memory. The retrieved events are then re-focused and combined
with the queries, the character embedding, and the world state to
form a \emph{mental posterior attributed to the actor.}

\subsection{Character Embedding}

To utilise all information presented in the history of the actor,
we encode $\tau_{j}$ to forward hidden states ($\overrightarrow{h}_{j}^{t}$)
and backward hidden states ($\overleftarrow{h}_{j}^{t}$) using the
bidirectional long-short term memory (Bi-LSTM) 

\begin{align}
\overrightarrow{h}_{j}^{t} & =\text{LSTM}_{\rightarrow}\left(\left[s_{j}^{t},a_{j}^{t}\right],\overrightarrow{h}_{j}^{t-1}\right),\label{eq:char-forward}\\
\overleftarrow{h}_{j}^{t} & =\text{LSTM}_{\leftarrow}\left(\left[s_{j}^{t},a_{j}^{t}\right],\overleftarrow{h}_{j}^{t+1}\right).\label{eq:char-backward}
\end{align}
By this structure, at time step $t$, the observer can retain and
reason about both the past and the future. Two sequences of forward
and backward states are employed as augmented information for the
value of the memory, as will be described in the next section.

Since the character of the actor is constructed by the history, we
summarised the last hidden state of the forward $\text{\text{LSTM}}$
over all past trajectories into the \emph{character embedding} of
the actor as
\begin{equation}
e_{char}=\frac{1}{N_{past}}\sum_{j=1}^{N_{past}}\text{ReLU\ensuremath{\left(\text{MLP}\left(\overrightarrow{h}_{j}^{T_{j}}\right)\right)}}.\label{eq:char-embedding}
\end{equation}
This character $e_{char}$ is supposed to govern the current behaviour
of the actor. To realise this, we treat the character embedding as
an additional input for another LSTM that processes the current unfinished
trajectory as follows:
\begin{equation}
h_{q}^{t}=\text{LSTM}\left(\text{concat}\left(e_{char},\left[s_{q}^{t},a_{q}^{t}\right]\right),h_{q}^{t-1}\right).\label{eq:mental-lstm}
\end{equation}
These LSTM states will later serve as raw materials for attributing
the mental states to the actor at each time step.

\subsection{Selective Attention}

The information extracted from $N_{past}$ past trajectories are first
stored in the key-value memory module $\mathcal{M}=\left\{ \left(k_{j}^{t},v_{j}^{t}\right)\right\} $
for $j=1,\dots,N_{past}$, and $t=1,\dots,T_{j}$. The memory \emph{key}
is a function of the forward state $k_{j}^{t}=g\left(\overrightarrow{h}_{j}^{t}\right)$.
The associated \emph{value} $v_{j}^{t}$ can be either (a) the forward
state (which contains \emph{actual} future information), or (b) the
concatenation of the forward state and the backward state or other
information computed from the rest of the trajectory from time $t+1$
(which contains \emph{predictive} information of the future). Let's
denote $\mathcal{M}.key=\left\{ k_{j}^{t}\right\} _{j=1\dots N_{past},}^{t=1\dots T_{j}}$
and $\mathcal{M}.value=\left\{ v_{j}^{t}\right\} _{j=1\dots N_{past}}^{t=1\dots T_{j}}$
is the set of all keys and the set of all values in the memory, respectively.

We then construct $M$ \emph{queries} to read out from the memory
based on selective events in the current unfinished trajectory. Let
$z_{q}^{T_{q}}=\text{MLP}\left(s_{q}^{T_{q}}\right)$ be the embedding
of the current world state. Given the set of hidden states $\mathcal{H}=\left\{ h_{q}^{t}\right\} ^{t=1\dots T_{q}}$
of all observable events in the current trajectory computed in Eq.~(\ref{eq:mental-lstm}),
we collect from this set $M$ selective events that are most similar
to $z_{q}^{T_{q}}$ as queries $\left\{ q_{m}\right\} $ for $m=1...M$
by using $\text{cosine}$ similarity, i.e. $\left\{ q_{m}\right\} _{m=1...M}=\left\{ h_{q}^{t'}\mid h_{q}^{t'}\in\mathcal{H},d_{z^{T_{q}}h_{q}^{t'}}\in\text{top-M}_{h_{q}^{t}\in\mathcal{H}}\left(d_{z^{T_{q}}h_{q}^{t}}\right)\right\} $
with $d_{z^{T_{q}}h_{q}^{t}}=\text{cosine}\left(z_{q}^{T_{q}},h_{q}^{t}\right)$.
Here, $\text{top-M}_{h_{q}^{t}\in\mathcal{H}}\left(d_{z^{T_{q}}h_{q}^{t}}\right)$
is a function that returns the set of $M$ highest values $d_{z^{T_{q}}h_{q}^{t}}$
given the set $\mathcal{H}$. The read head uses the queries in parallel
to retrieve memory content as: 
\begin{equation}
\bar{v}_{m}=\sum_{v_{j}^{t}\in\mathcal{M}.value}\text{attn}\left(q_{m},k_{j}^{t}\right)v_{j}^{t},\label{eq:retrieved-mem}
\end{equation}
where $\text{attn}\left(q_{m},k_{j}^{t}\right)$ is the soft attention
score which is computed as 
\begin{equation}
\text{attn}\left(q_{m},k_{j}^{t}\right)=\frac{e^{d_{mjt}/\beta}}{\sum_{k_{j'}^{t'}\in\mathcal{M}.key}e^{d_{mj't'}/\beta}},\label{eq:attention}
\end{equation}
with the temperature $\beta>0$ and the distance $d_{mjt}=\text{cosine}\left(q_{m},k_{j}^{t}\right)$.

\subsection{Mental Attribution}

\paragraph{Re-focusing on selective events}

The $\Model$ does not treat all the selective events equally, instead,
it will re-weight these selective events based on the embedding of
the current world state $z_{q}^{T_{q}}$ and $\left\{ q_{m}\right\} $
via a set of \emph{attention weights}

\[
\alpha_{m}=\frac{e^{\delta_{m}/\beta}}{\sum_{m'=1\dots M}e^{\delta_{m'}/\beta}},\text{ for }m=1\dots M
\]
where $\delta_{m}$ is a metric measuring the relationship between
a recent event $z_{q}^{T_{q}}$ and the selective event $q_{m}$.
The distance $\delta_{m}$ is generated by a neural network $\text{MLP}\left(\left[z_{q}^{T_{q}},q_{m}\right]\right)$
which attempts to learn a metric to measure the importance of a selective
event $q_{m}$ to the behaviour predictions made at the recent event.
This mechanism captures the re-focusing process on a smaller and more
selective set of events in the trajectory. 

\paragraph{Mental Posterior}

Based on the attention weights, the retrieved memory contents, combined
with the queries, constitute the current mental state of the actor:
\[
e_{mental}=\frac{1}{M}\sum_{m=1}^{M}\alpha_{m}\text{concat}\left(\bar{v}_{m},q_{m}\right).
\]
This mental state, together with the last hidden state $h_{q}^{T_{q}-1}$,
the character embedding $e_{char}$ in Eq\@.~(\ref{eq:char-embedding})
and the representation of the current state $z_{q}^{T_{q}}$, form
the \emph{mental posterior} $\mathbf{e}_{p}=\text{concat}\left(e_{mental},h_{q}^{T_{q}-1},e_{char},z_{q}^{T_{q}}\right)$
that serves as input for the prediction heads. 

We use four prediction heads for predicting preference (or goal),
one-step ahead intention, one-step ahead action, and the successor
representations.

\subsection{Training}

Our neural network is trained by minimising a combined loss for preference
(or goal) prediction ($\mathcal{L}_{\text{pref}}$), intention prediction
($\mathcal{L}_{\text{intent}}$), action prediction ($\mathcal{L}_{\text{action}}$),
and the successor representations prediction ($\mathcal{L}_{\text{SR}}$)
as

\[
\mathcal{L}=\mathcal{L}_{\text{pref}}+\mathcal{L}_{\text{intent}}+\mathcal{L}_{\text{action}}+\mathcal{L}_{\text{SR}}.
\]
The first three component losses are negative log-likelihoods of the
corresponding targets and are computed as follows:
\[
\mathcal{L}_{\text{pref}}=\sum_{\text{pref}}-\log P\left(\text{pref}\left|\mathbf{e}_{p}\right.\right),
\]
\[
\mathcal{L}_{\text{intent}}=\sum_{\text{intent}}-\log P\left(\text{intent}\left|\mathbf{e}_{p}\right.\right),
\]
\[
\mathcal{L}_{\text{action}}=-\log P\left(a_{t}\left|\mathbf{e}_{p}\right.\right)
\]
where the $\mathbf{e}_{p}$ is the mental posterior. To compute the
successor representation loss $\mathcal{L}_{\text{SR}}$, we first
compute the empirical successor representation (SRs) by $SR_{\gamma}^{(t)}\left(s\right)=\frac{1}{Z_{t}}\sum_{t'=0}^{T-t}\gamma_{SR}^{t'}I\left(s_{t+t'}=s\right)$
where $T$ is the episode length, $t$ is the time at which the successor
representation is computed, $\gamma_{SR}\in\left(0,1\right)$ is the
discount factor, $Z_{t}=\sum_{s\in\mathcal{S}}\sum_{t'=0}^{T-t}\gamma_{SR}^{t'}I\left(s_{t+t'}=s\right)$
is a normalisation constant ($\mathcal{S}$ is the state space), and
$I\left(s_{t+t'}=s\right)$ is an indicator function, which returns
$1$ if $s_{t+t'}=s$ and $0$ otherwise. We then use the cross-entropy
loss to obtain the SR loss

\[
\mathcal{L}_{\text{SR}}=\sum_{\gamma_{SR}}\sum_{s}-SR_{\gamma_{SR}}^{(t)}\left(s\right)\log\widetilde{SR}_{\gamma_{SR}}^{(t)}\left(s\right).
\]

By training, $\Model$ learns a \emph{prior model} of others from
observing and predicting actors' behaviours, which is captured in
network weights and the analogy-making capability. When mentalising
about an actor, $\Model$ updates the posterior upon seeing some of
its behaviours.

\section{Experiment Results}

We evaluate $\Model$ on multiple tasks, including predicting preference,
intention, action, and successor representations as well as assessing
false belief understanding. For simplicity, we set the memory keys
to the forward LSTM states ($k_{j}^{t}=\overrightarrow{h}_{j}^{t}$
of Eq.~(\ref{eq:char-forward})). Similarly, the memory values are
set to either the forward LSTM states ($v_{j}^{t}=\overrightarrow{h}_{j}^{t}$
of Eq.~(\ref{eq:char-forward})), or the concatenation of both forward
and backward LSTM states ($v_{j}^{t}=\left[\overrightarrow{h}_{j}^{t},\overleftarrow{h}_{j}^{t}\right]$
of Eq.~(\ref{eq:char-forward}) and Eq.~(\ref{eq:char-backward})).
The latter is called \texttt{Bi-}$\Model$. The number of queries
is set as $M=10$.

In practice, some actions (such as \texttt{\small{}pick-up}) happen
far less frequently than others in the whole sequence. Thus we use
a replay buffer to balance the class of actions in training. As the
relay buffer plays the role of episodic memory in the learning process,
we call this balancing strategy action-based episodic memory (AEM).
For comparison, we implemented a recent representative neural ToM
network called ToMnet \cite{rabinowitz2018machine}. 

\subsection{Light-Room Environment}

\begin{figure}
\begin{centering}
\includegraphics[width=0.8\columnwidth]{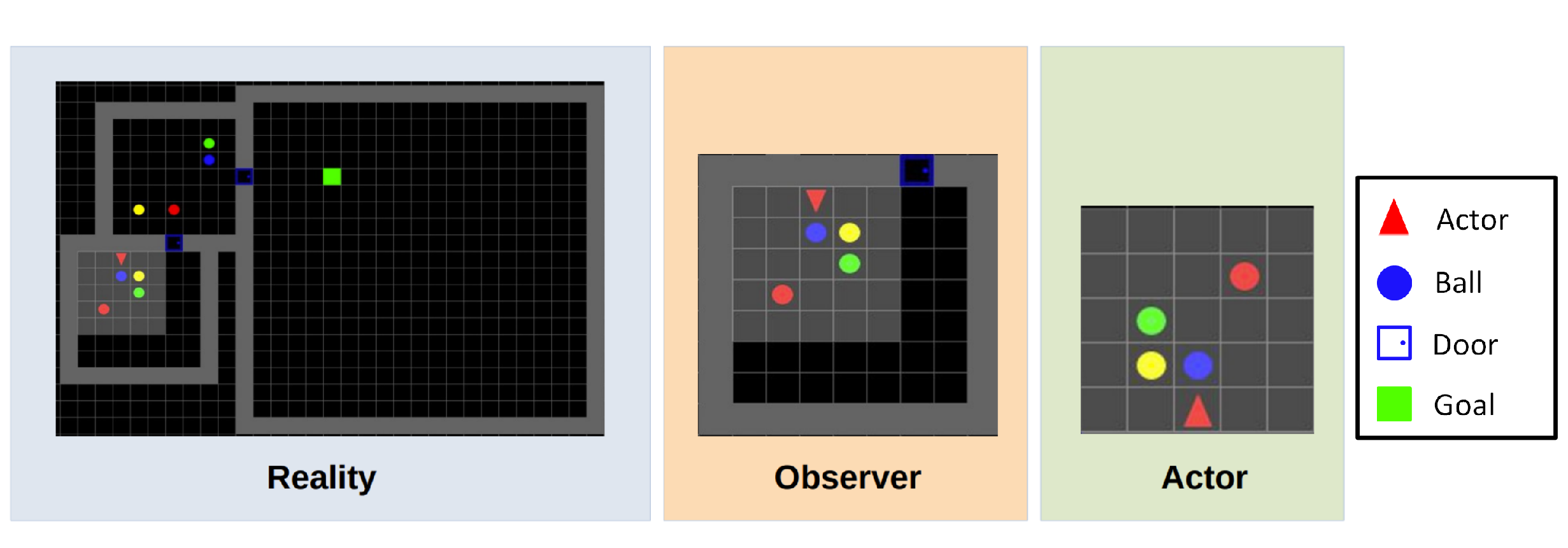}
\par\end{centering}
\caption{\label{fig_lightroom} A multi-light-room environment with three adjoining
rooms. The observer has a different perspective from the actor, and
both can only partially observe reality.}

\end{figure}

To study ToM models, we created a multi-light-room environment using
the \emph{gym-minigrid} framework \cite{gym_minigrid} (see Fig.~\ref{fig_lightroom}).
The observer (ToM agent) can only see the lit room where the actor
is in. After the actor leaves a room, the light will be turned off,
and the observer will not see what happens in this room afterwards,
e.g. swapping keys. As a result, the observer needs to memorise what
happened in all rooms to read the current mind of the actor and to
predict the actor's behaviours correctly.

We procedurally generated the actor's behaviours as follows. At each
step, the actor chooses one amongst three intentions \texttt{\small{}find()},
\texttt{\small{}goto()}, \texttt{\small{}pickup()} and executes the
intention by choosing between four primitive actions \{\texttt{\small{}turn-left},
\texttt{\small{}turn-right}, \texttt{\small{}move-forward}, \texttt{\small{}pickup}\}.
To find an object, the actor first hypothesises an arbitrary position
in the room and then walks to this position to verify. If the actor
could not find the object, it will make another hypothesis. We call
this type of actor a \emph{hypo-actor}. After seeing any object, the
actor will hold a belief about the position of this object. This belief
can be changed if the actor recognises that the object no longer exists
in the original position. Each actor can have a small field of view
as $3\times3$, e.g. it can observe a square of $3\times3$ in front
of it or has a larger field of view as $5\times5$ (the left-most
figure of Fig.~\ref{fig_lightroom}).\\

\subsection{Preference, Action and Intention Prediction}

We carry out experiments where the observer uses its knowledge about
the past and current trajectories of the actor to predict the actor's
preference, actions and intentions. In these settings, the actor has
its own preference for the ball and tries to navigate to collect the
ball in its preferred colour in the first light-room. It will try
to navigate to the goal located in the final room. In other light-rooms
between the first and the final room, balls of different colours exist.
Since the actor has already picked up the preferred ball, it will
no longer pick up other balls; therefore, these other balls can be
considered as distractors to the observer. We trained ToM models in
episodes with three light-rooms and test models under different conditions:
(1) past trajectories are in three light-rooms, and the current trajectory
is in three light-rooms; (2) past trajectories are in three light-rooms,
and the current trajectory is in five light-rooms. 

\subsubsection{Preference Prediction}

\begin{figure}
\begin{centering}
\includegraphics[width=0.85\columnwidth]{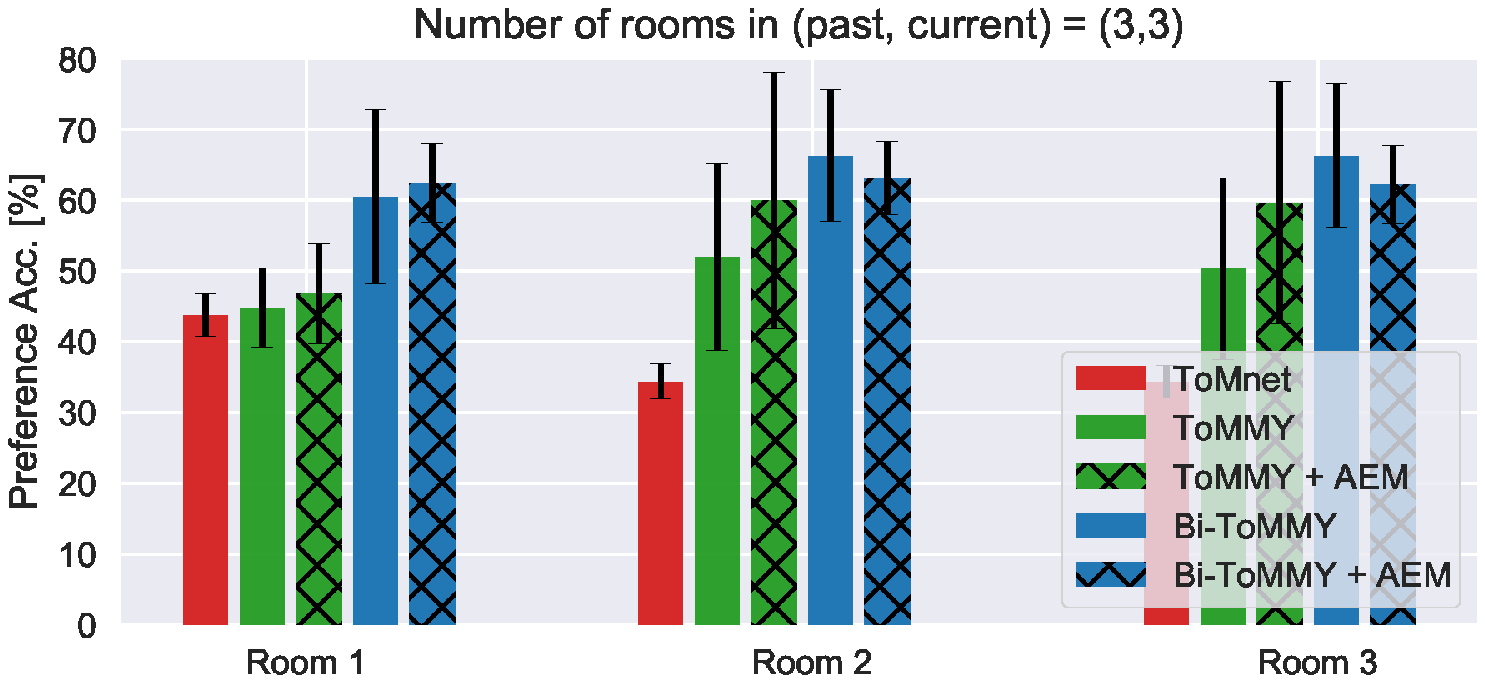}
\par\end{centering}
\begin{centering}
\includegraphics[width=0.85\columnwidth]{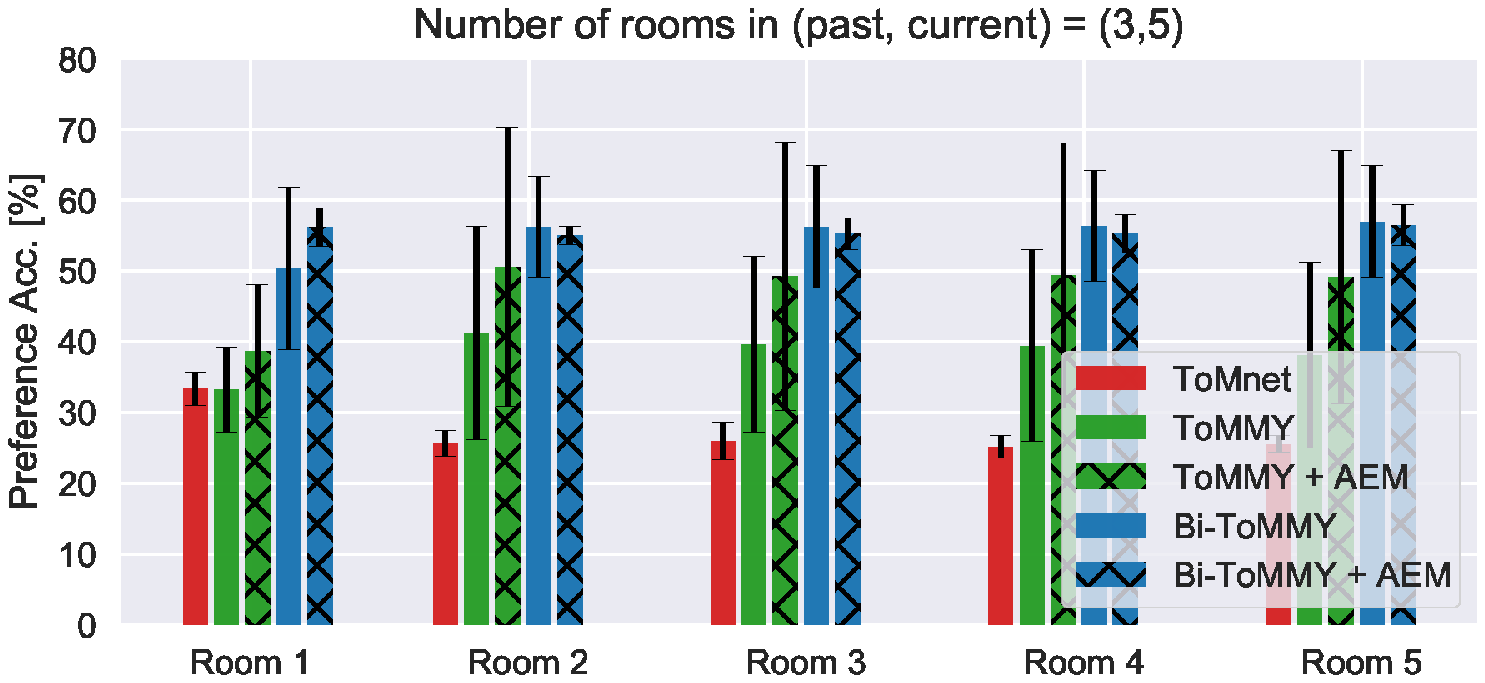}
\par\end{centering}
\caption{\label{fig_pref_pred} Preference prediction (mean and std.) of ToMnet
and $\protect\Model$s across rooms in two scenarios: (top) both current
and past trajectories have three rooms; and (bottom) there are three
and five rooms in the past and current trajectories, respectively.
The performance is measured when the agent presents in each room during
the current trajectory. }
\end{figure}

Fig.~\ref{fig_pref_pred} shows the performance of ToMnet \cite{rabinowitz2018machine}
and $\Model$s on preference prediction tasks. The mean and standard
deviation (std.) are computed over 4 runs for each model. Since ToMnet
uses LSTM to compress the entire history of the actor into a single
character embedding vector, it will struggle to remember details of
the long past, which are critical to giving the correct answer in
this situation. In our experiment, $\Model$s give better answers
than the ToMnet during the episode. This is because $\Model$s effectively
querying past trajectories by the memory mechanisms. 

\subsubsection{Action and Intention Prediction}
\begin{figure}
\begin{centering}
\includegraphics[width=0.95\columnwidth]{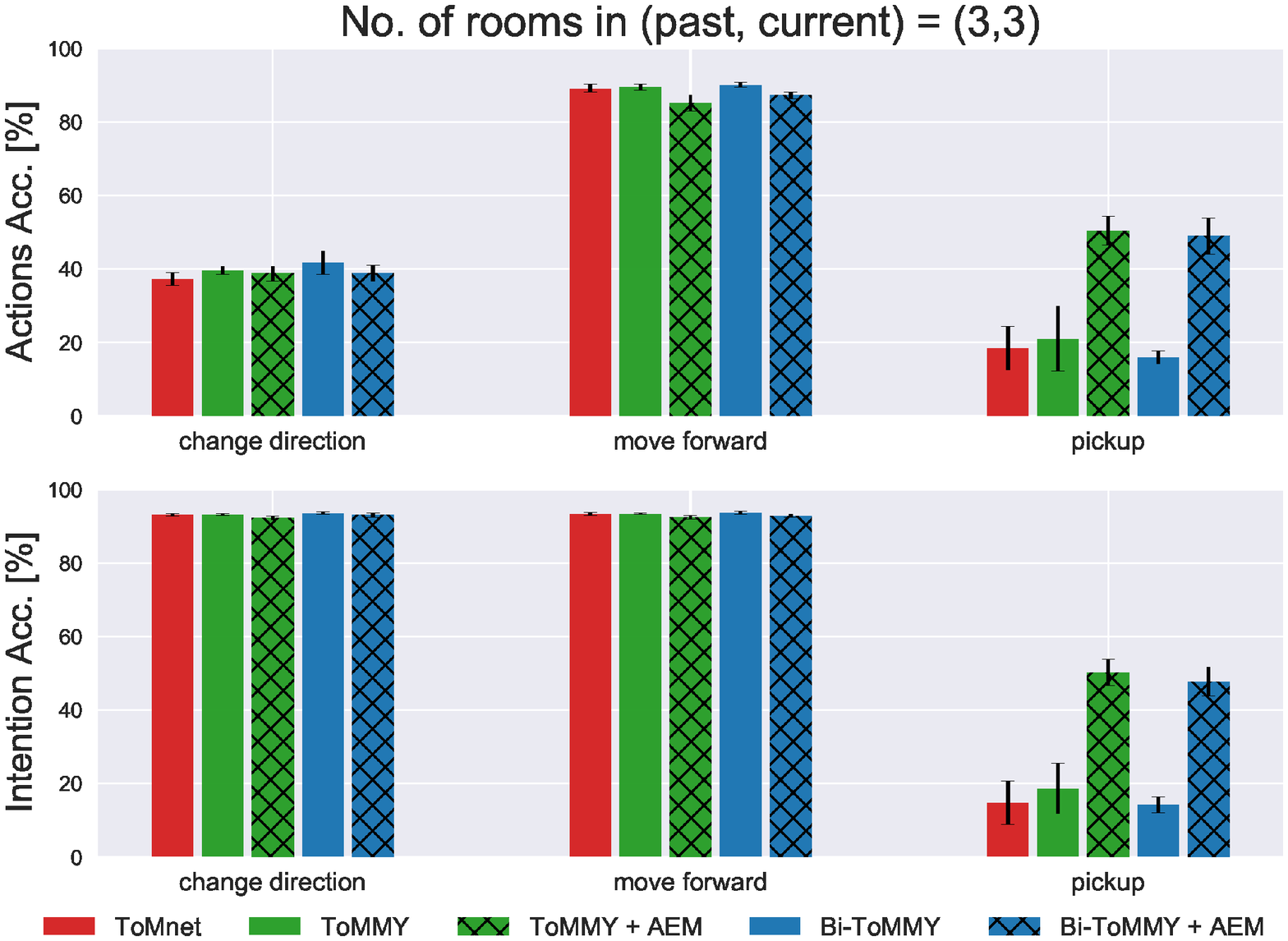}
\par\end{centering}
\begin{centering}
\includegraphics[width=0.95\columnwidth]{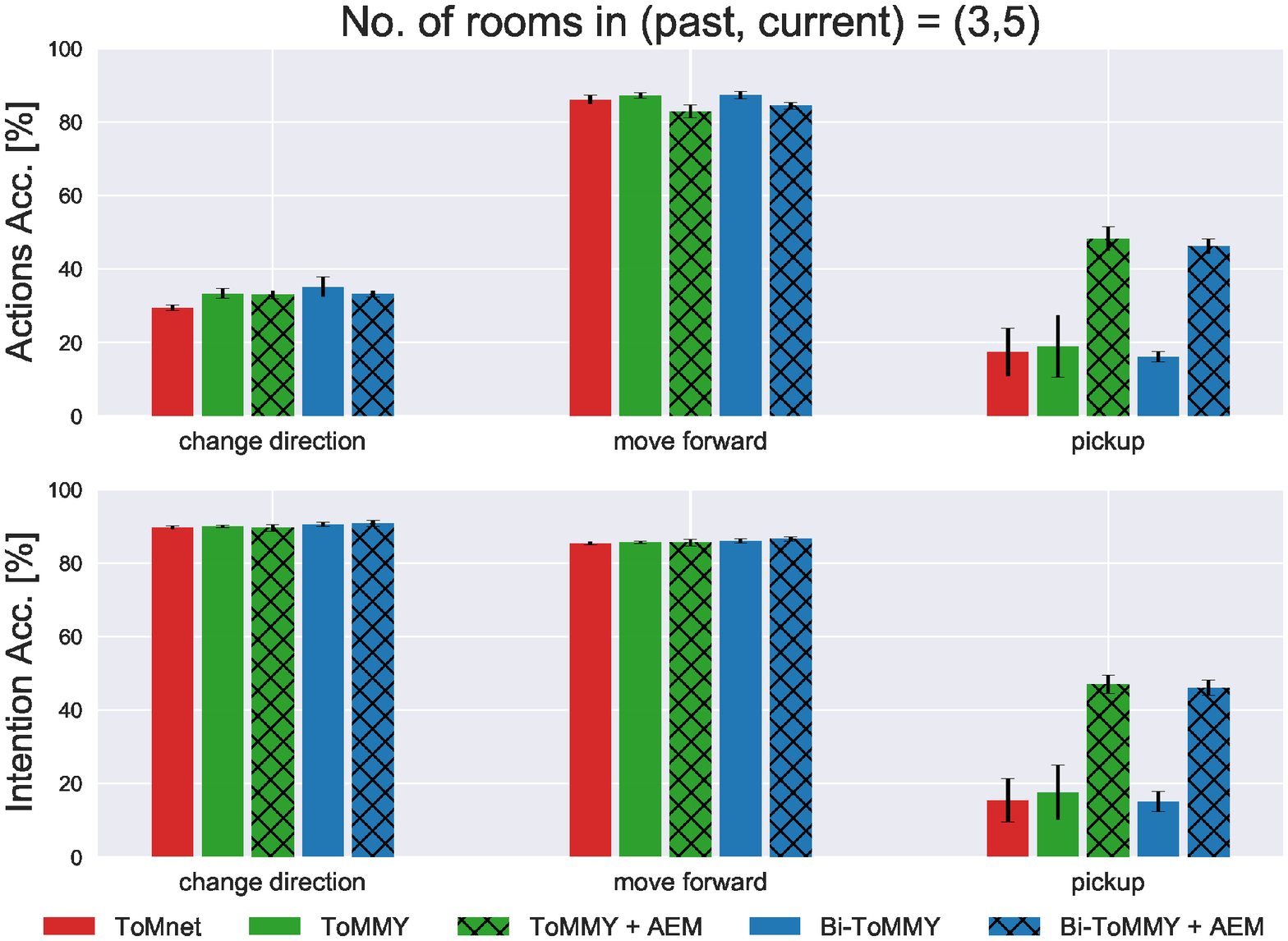}
\par\end{centering}
\caption{\label{fig_action_intent_pred} Performance of ToMnet and $\protect\Model$s
on the action and intention prediction (mean and std.). The x-axis
shows three groups of actions: (1) change direction or \texttt{\small{}(turn-left}
or \texttt{\small{}turn-right}), (2) \texttt{\small{}move-forward},
and (3) \texttt{\small{}pickup}. }
\end{figure}
\begin{figure}
\begin{centering}
\includegraphics[width=1\columnwidth]{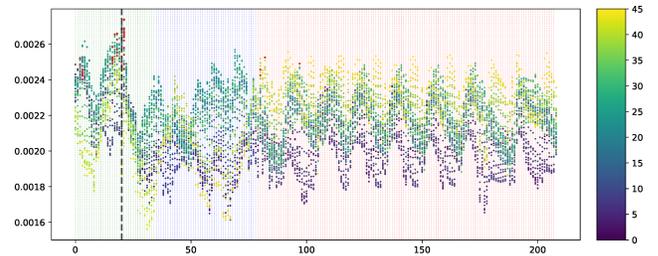}
\par\end{centering}
\caption{\label{fig_weights_past_traj} Visualisation of the reading weights
from the beginning of the current trajectory up to the moment the
actor picks up the ball (here is at step $45$). The heatmap colour
indicates the time in the current trajectory at which the query is
conducted. The red dots are the peak weights of trajectories. The
background colours indicate the rooms that the actor was in. Rooms
1, 2 and 3 are coded as \textcolor{green}{green}, \textcolor{blue}{blue},
and \textcolor{red}{red}. The vertical black dash line indicates when
the actor picked up the ball in the past trajectory. Our query mechanism
generates the weights with high values when the actor is in room 1,
especially when it picks up the ball, e.g. high peak at the vertical
black dash line.}
\end{figure}

Fig.~\ref{fig_action_intent_pred} shows the performance of ToM models
on action and prediction tasks. Since $\Model$s can efficiently use
past information, it can predict better over ToMnet when the actor
changes its direction to look for an object. The models that used
bidirectional long-short term memory to process the past trajectory
(\texttt{Bi-}$\Model$) can improve the performance in action prediction
when the actor changes its direction. In all settings, when the actor
picks up an object--a rare event--only the methods augmented with
the action-based episodic memory can learn to predict correctly.

\subsubsection{Visualisation}

Fig.~\ref{fig_weights_past_traj} shows the attention weights (in
Eq\@.~(\ref{eq:attention})) over experiences in past trajectories.
The attention weights generated by the network are relatively higher
during the period when the actor was in the first room, especially
when it picked up the ball. This means the ToM agent learnt to correctly
attends to moments that express the actor's preference. 

\subsection{High Demand False-belief Assessment}

\begin{figure}
\begin{centering}
\includegraphics[width=0.95\columnwidth]{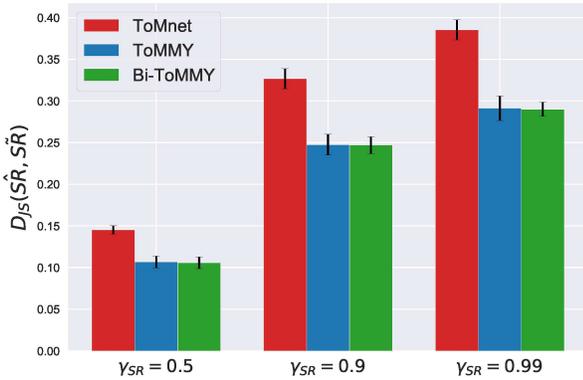}
\par\end{centering}
\caption{\label{fig_djs_high_demand_false_belief} Performance of theory of
mind agents (ToMnet, $\protect\Model$ and Bi-$\protect\Model$) in
high demand false-belief task measured by the Jensen-Shannon divergence
(mean and std.) between the successor representations ($\gamma_{SR}=0.5$,
$\gamma_{SR}=0.9$, $\gamma_{SR}=0.99$) predicted by the models and
the ground truth (the lower the better). }
\end{figure}

\begin{figure*}
\centering{}\includegraphics[width=0.95\textwidth]{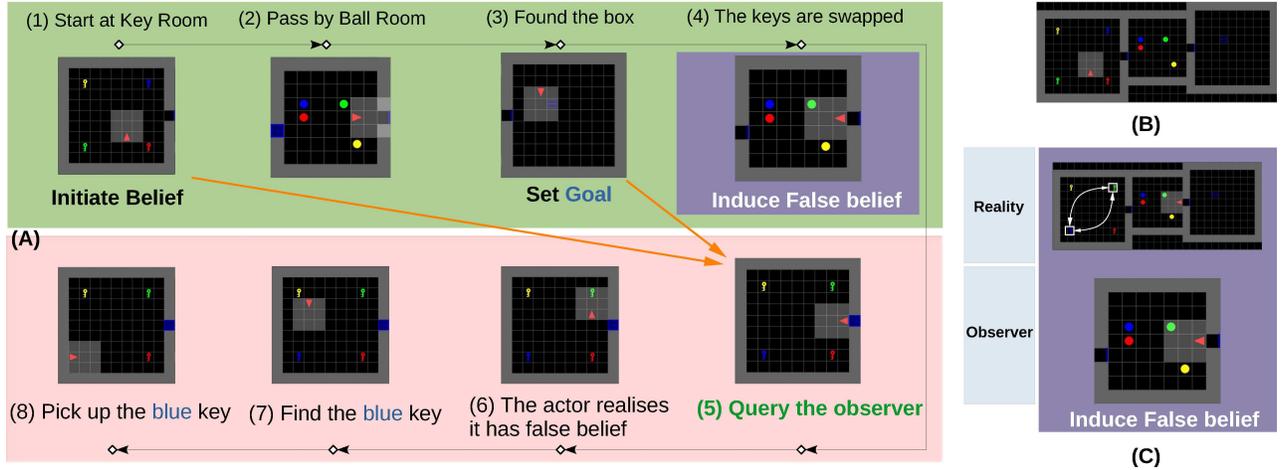}\caption{\label{fig_illu_lightroom_high_false_belief} High demand false-belief
task in light-room environment. (A) The sequence of observations given
to the theory of mind agent (from stage 1 to 5) and the subsequence
events in the \textbf{key room }when the keys are swapped (stage 6
to stage 8); (B) The full observation of the environment and the reality
at the initial state; (C) The moment when keys are swapped. At stage
5, to predict the successor representations, the observer must remember
the actor's goal that is revealed at stage 3 and recall the positions
of the keys seen at the beginning of the episode/stage 1 (two \textcolor{orange}{orange}
arrows).}
\end{figure*}

We constructed a three-light-room environment that demands a high
mental load for theory of mind. This setting is inspired by the emerging
human-machine teaming scenarios in which a ToM-equipped robot companion
follows a human actor. This robot has the privilege of seeing more
than what the human can see in the current room, but may not know
what happens in the other room, e.g. the keys are swapped unseen
by the observer. Hence, the robot needs to remember the past well
to recognise human's false beliefs and then uses this information
to provide proper help. A scenario is illustrated in Fig.~\ref{fig_illu_lightroom_high_false_belief}.

In this setting, there are three rooms: (1) The \textbf{key room},
(2) the \textbf{ball room} as distractors, and (3) the \textbf{box
room} where the goal is revealed. The actor has an initiated belief
about the position of keys located in the \textbf{key room} (stage
1). The actor starts from the \textbf{key room}, passes by the \textbf{ball
room} (stage 2), and arrives in the \textbf{box room}. Here the actor
finds the box whose colour matches with that of the key which the
actor should collect (stage 3). The actor then comes back to the \textbf{key
room} with the goal is to collect the right key. When the actor passes
by the \textbf{box room} the second time, the position of the keys
can be changed without the actor knows. This induces false beliefs
in the actor (stage 4). When the actor comes back to the \textbf{key
room} the second time, we ask the observer to predict the successor
representations of the actor (stage 5). At this stage, the observer
does not have false beliefs since it can observe the current position
of the keys. However, to know that the actor has false beliefs, the
observer needs to refer to the position of the keys in the \textbf{key
room} at the beginning of the episode. Also, to infer the actor's
goal, the observer needs to recall the event in the \textbf{box room}.

In this test, the past trajectories do not provide any information
about the goal or preference like in previous experiments but reveal
the behaviour of the actors. For example, it would help the observer
predict the distance the actor can see the object. The current trajectory
contains information about the actor's goal in the middle of the trajectory.
This setting challenges the ToM models without the ability to recall
the key's position to predict the actor's behaviours. 

\paragraph{Results}

$\Model$s predict more accurately the successor representations at
the time the actor comes back to the key room, as shown in Fig.~\ref{fig_djs_high_demand_false_belief}.
At this moment, to answer correctly, the theory of mind agents must
recall the position of keys at the beginning of the episode when the
actor was at the key room the \emph{first time} and the colour of
the box in the box room. These important events are divided twice
times by periods when the actor is in the distractor rooms. Failing
to recall this information will lead to incorrect predictions. 

\paragraph{Visualisation}

\begin{figure*}
\begin{centering}
\subfloat[Weak Correction]{\begin{centering}
\includegraphics[width=1\textwidth]{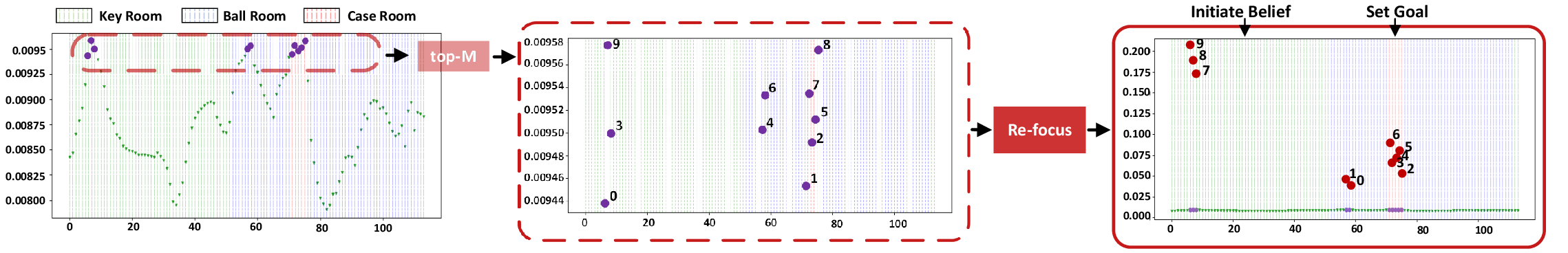}
\par\end{centering}
}
\par\end{centering}
\begin{centering}
\subfloat[Strong Correction]{\begin{centering}
\includegraphics[width=1\textwidth]{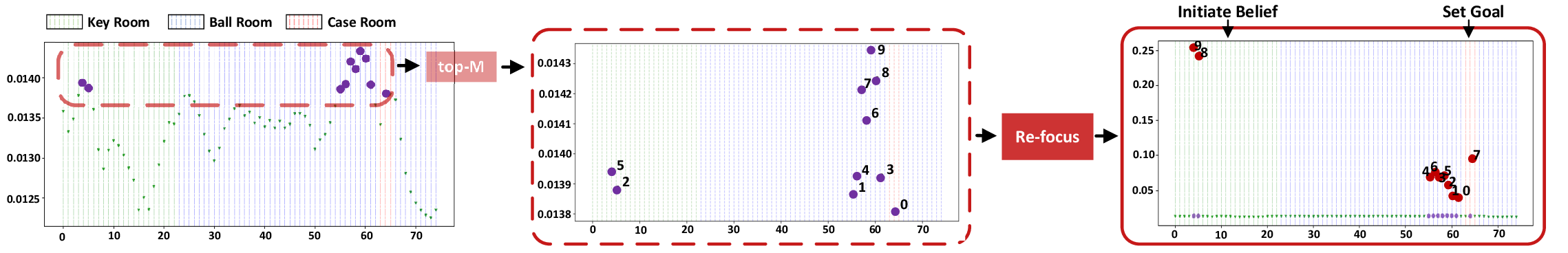}
\par\end{centering}
}
\par\end{centering}
\caption{\label{fig_weights_current_traj_high_false_belief} Each figure shows
the attention weights over the current trajectory right after the
actor returns to the key room. The key room, ball room, box room are
coded as \textcolor{green}{green}, \textcolor{blue}{blue}, and \textcolor{red}{red},
respectively. The purple circles indicate the top ten attention weights.
The red circles indicate the top-M attention weights after re-focusing.
The number over the circles shows the rank of the weights magnitude,
i.e., the higher number, the more critical. When asked to predict
future behaviour, the observer recalls when the actor was in the key
room (to the original position of the keys) and box room (the actor
reveals its goal). Figures in the left-most column are the attention
weights generated by the cosine similarity. The middle column is a
closer look at the top-M selective events $\left\{ q_{m}\right\} _{m=1\dots M}$.
The right-most column presented the weights over top-M selective events
reshaped by the re-focusing mechanism $\left\{ \alpha_{m}\right\} _{m=1\dots M}$.
In the first case (a), the re-focusing mechanism only needs to provide
weak correction since events that have the highest attention weights
after the cosine similarity are events when the actor is in the box
room and the key room. However, in the second case (b), the strong
correction from the re-focusing mechanism is important.}
\end{figure*}

Fig.~\ref{fig_weights_current_traj_high_false_belief} shows that
$\Model$ attends to the period in the current trajectory that the
actor is in the box room when the goal is revealed. It also refers
to the beginning of the episode to recall the keys' original position
to know whether keys were swapped. In case the weights according to
top-M selective events are not highlighted on events where the actor
was in the key room and the box room, the model is able to re-generate
weights based on this small set of events. As shown in Figure \ref{fig_weights_current_traj_high_false_belief}.a,
although the two highest weights which are generated by the cosine
similarity metric are on important events when agents are in the box
room and the key room, the re-focusing mechanism still gently corrects
the attention of $\Model$ by decreasing the weights of other events
in the distractor room. Especially, when the attention over top-M
selective events highly rises up in the distractor events, as in Figure
\ref{fig_weights_current_traj_high_false_belief}.b, this mechanism
is crucial to correct and help the theory of mind agent re-focuses
on the important events. Hence, $\Model$ understands whether the
actor may have false beliefs.

\section{Conclusion}

Aiming at equipping artificial agents with new social capacities we
introduced $\Model$, a new neural theory of mind model that utilises
the power of external memory and hierarchical attention for mentalising
over complex behaviours of other agents in POMDPs settings. The memory
facilitates meta-learning from prior experiences the analogy-making
capability in social situations without the need of explicit domain
knowledge or task structures. This capability is then refined when
$\Model$ sees an actor and its past and current behaviours. We also
introduced a new high-demand false-belief task to assess the theory
of mind ability to understand if others wrongly believe in things
that no longer hold. Our experiments showed that memory facilitates
the learning process and achieves better social understanding, especially
in theory of mind tasks that demand a high cognitive load. 
\bibliographystyle{aaai23}
\bibliography{main}

\end{document}